\documentclass[letterpaper]{article}
\usepackage{natbib,alifeconf,hyperref, xcolor}   
\hypersetup{backref, colorlinks=true, 
linkcolor=red, 
linkbordercolor = red,          
citecolor = blue}

\title{Chemlambda, universality and self-multiplication}
\author {Marius Buliga $^{1}$ \and Louis H. Kauffman $^{2}$\\
      \mbox{} \\
      $^1$ Institute of Matematics of the Romanian Academy \\ P.O. BOX 1-764, RO 014700, Bucharest, Romania \\ 
\texttt{Marius.Buliga@gmail.com} \\
     $^2$ Department of Mathematics, University of Illinois at Chicago \\ 851 South Morgan Street, Chicago, Illinois, 60607-7045 \\  \texttt{kauffman@uic.edu} \\
 }

\begin{document}
\maketitle

\begin{abstract}
We present chemlambda (or the chemical {\em concrete} machine), an artificial chemistry with the following properties: (a) is Turing complete, (b) has a model of decentralized, distributed computing associated to it, (c) works at the level of individual (artificial) molecules, subject of reversible, but otherwise deterministic interactions with a small number of enzymes, (d) encodes information in the geometrical structure of the molecules and not in their numbers, (e) all interactions are purely local in space and time. This is part of a larger project to create computing, artificial chemistry and artificial life 
in a distributed context, using topological and graphical languages.
\end{abstract}

\section{Introduction}
\label{proper}

In this note we want to briefly present chemlambda (or the chemical {\em concrete} machine), an artificial chemistry with the following properties: (a) is Turing complete, (b) has a model of decentralized, distributed computing associated to it \citep{Web}, (c) works at the level of individual (artificial) molecules, subject of reversible, but otherwise deterministic interactions with a small number of enzymes, (d) encodes information in the geometrical structure of the molecules and not in their numbers, (e) all interactions are purely local in space and time. 

In some respects chemlambda is closed to the fraglets \citep{tschudin1} and metabolic approaches \citep{tschudin2} research line.  In others, it resembles to the CHAM ("chemical abstract machine")  \citep{bb}, which  uses a chemical metaphor for modeling  asynchronous concurrent computations (in particular a concurrent lambda calculus).  Algorithmic Chemistry  \citep{fonbas1} \citep{fonbas2} \citep{fonbas3} is another classical line of inspiration. Because it concentrates at the level of individual molecules, it departs however from the programming model of computation  introduced in  \citep{banatre2} \citep{banatre1}.

Chemlambda appeared as an artificial chemistry version of a graph rewrite system, called graphic lambda calculus (GLC) \citep{bgraph} \href{http://chorasimilarity.wordpress.com/graphic-lambda-calculus/}{(web tutorial)}. In GLC programs are certain trivalent graphs, and execution of programs means the application of graph rewrites, called "moves", on the respective graph. 

In the GLC formalism there is one global move (GLOBAL FAN-OUT), all the other moves are local (i.e. they involve a fixed, small number of nodes).

Chemlambda  was introduced in order to eliminate this unpleasant GLOBAL FAN-OUT. Chemlambda  uses only local moves \citep{buligachem} \href{http://chorasimilarity.wordpress.com/chemical-concrete-machine/}{(web tutorial)}. The moves of chemlambda act on trivalent graphs called "molecules" at certain "reaction sites", like chemical reactions involving molecules and enzymes (here  enzyme=move).

Later on, a distributed, decentralized model of computation appeared, called distributed GLC \citep{Web}, which is based on chemlambda and GLC, also using the Actor Model by Hewitt   \citep{hewitt2} \citep{actor}. The Actor Model ingredient is  a replacement of proximity relations between (individual) interacting molecules. Indeed,  real chemical interactions happen only  between  molecules which are close one to another. But in the chemlambda formalism there is no space where these artificial molecules float, they are just certain trivalent graphs, not embedded in any way in a space. Computation with chemlambda molecules is seen as asynchronous, purely local and decentralized application of graph rewrites (i.e. moves, or interaction with enzymes). Proximity relations are then replaced by interactions between actors, each actor being in charge of a molecule, and having a very limited repertoire of behaviours. (In turn, each behaviour uses one of the graph rewrites available, either applied between two interacting actors, or internally, as it is the case of the sequences of moves which effect a self-multiplication replacing the GLOBAL FAN-OUT of GLC).

The key merit of this model is a graphical reformulation of the well-known lambda calculus, central to logic and to the design of recursion in computer languages. By reformulating the lambda calculus in terms of graphs, the operations for the calculus become essentially local operations of graphical replacement. The graphs themselves contain all the data that is usually formulated in terms of algebra. This means that the global structure of the graph contains all the information that is usually cut up into bits of algebra. The graph becomes a whole system that instantiates the computational power of the calculus. This instantiation is the key reason why this model can propose significant
designs in distributed computing. The graph as a whole can exist in a widely distributed fashion, while the interactions that constitute its computations are controlled by local nodal exchanges between actors. 

Furthermore, this property of redesigning the relationship of the local and the global is
not restricted just to lambda calculus networks. There are relationships
of the same kind that link this research with topology (knots and lambda calculus  \citep{kauffman2}, knot automata  \citep{kauffman3}), or with topological quantum computing \citep{chen}, \citep{kauffman1}.

\section{A quick review of lambda calculus}
In this section we give a very quick review of the formalism and ideas of lambda calculus.
First of all the notation $$F = \lambda xy.f(x,y)$$ indicates a function $f(x,y)$ of two variables, defined in some domain and a stipulation (the part after $\lambda$ and before the function) of the order of application of the operator $F$ to these variables. Thus  we can write 
$$(Fx)y = f(x,y).$$ For example, If $$F = \lambda xy.y(yx),$$ then $$(Fa)b = b(ba).$$
Later we will make a notation for the operation of evaluating such an operator, but for now we just consider the non-associative algebra structure of such operators. We can work in reverse as well.
Suppose I say that $G$ is an operator defined by the equation $$((Gx)y)z = (yx)(yz).$$ Then we have in the $\lambda$ notation, $$G = \lambda xyz.(yx)(yz).$$
\bigbreak

For this analysis, let us suppose that the algebra generated by the variables $x,y,z,\cdots$ is a universal non-associative algebra. This means that the binary operation is non-associative and there are no further relations instantiated. However, if we define $M$ by $Mx = xx$ and regard $M$ as an element of an extension of the original algebra by giving it the status of $M=\lambda x.xx,$ then $M$ satisfies the special relation that defines it and furthermore we would like to be able to say that the definition of the action of $M$ applies even if we apply $M$ to itself. In that case we would have $MM = M$ as a consequence of the definition $M = \lambda x.xx.$ Thus we can start with a universal non-associative algebra and then add new elements that satisfy special relations. We can in this freely made situation
allow the new elements to act (compose) upon themselves. 
\bigbreak

Here is a useful example. Let $F$ be a given operator. It can be one of the original variables, or it  can be a defined operator such as we have discussed above. The we define $G$ as
$$Gx = F(xx).$$ That is, we define $$G = \lambda x.F(xx).$$ Now we note that 
$$GG= F(GG).$$ Thus any $F$ in our algebra has a fixed point that is another element of the algebra.
This is the Fixed Point Theorem of Church and Curry. Along with this fixed point theorem comes some 
caution in the use and construction of such lambda calculi. For suppose we had been dealing with a logical calculus and $F = \sim $, the negation operator. Then in our initial calculus we may have assumed that negation does not have a fixed point, as in classical logic. But we have seen that 
if $$G = \lambda x.\sim (xx),$$ then $$GG = \sim (GG).$$ Thus the extended algebra can not be expected to 
continue to obey classical logical rules. If it is desired to continue to obey such rules then one must put
some controls on the lambda calculus. Also, if one has a fixed point as in $$GG =F(GG),$$ then there is 
the possibility of an infinite recursion of the form
$$GG = F(GG) = F(F(GG)) = F(F(F(GG)))=$$ 
$$= F(F(F(F(GG))))=F(F(F(F(F(GG)))))= \cdots .$$
It is good to have a formalism for recursion, but the language needs to include controls for that so that a computation does not run without stopping.
\bigbreak

One way to handle such control is to replace equality of evaluation by an evaluation or reduction step.
Then one would have $$(\lambda x.H(x))a \longrightarrow H(a)$$ where the arrow refers to a reduction step that can be performed. In this case, the step is called {\it beta - reduction}. In the rest of this paper, we show how to adapt such controlled lambda calculi to operations on graphs where steps of replacement from one graph to another correspond to operations like beta-reduction. The graphs, once they are formulated, have the advantage that the details of labeling in the algebra have disappeared into graphical connections and so certain complexities of lambda calculi are handled automatically.
We envisage such graphical systems and their evolutions under computational steps such as beta-reduction as a new and powerful formulation of computation and information processing.
\bigbreak

\section{The Chemlambda formalism}

Chemlambda is a graph rewriting system. It consists in a family of graphs, called "molecules" and a list of graph rewrites, called "moves".  Every move is  local, in the sense that there is an a priori upper bound on the number of nodes and arrows which are modified during the move.

\begin{figure} [ht]
      
     \includegraphics[width=  75mm]{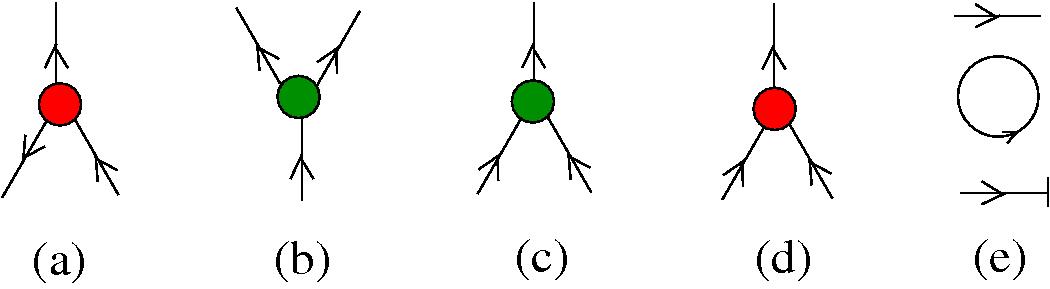}
     \caption{Basic pieces of chemlambda molecules: (a) lambda abstraction node, (b) fanout node, (c) application node, (d) fanin node, (e) arrow, loop and termination node}
     \label{4_chem}
 
\end{figure}

A {\em molecule} is a locally planar graph made by a finite number of pieces (arrows, loops, nodes described in Fig.~\ref{4_chem}). We may admit also a set of nodes with unspecified valences, called "other molecules". These are the equivalent of "cores" from \citep{Web} section 3,  paragraph 5. Interaction with cores, i.e. they can be used as interfaces with external constructs.

Each chemlambda {\em move} is to be interpreted as the interaction of a chemlambda molecule with an enzyme (which has the same name as the move), at a certain reaction site (the place where the move is applied). The moves are reversible. 

\begin{figure} [ht]
      
     \includegraphics[width=  75mm]{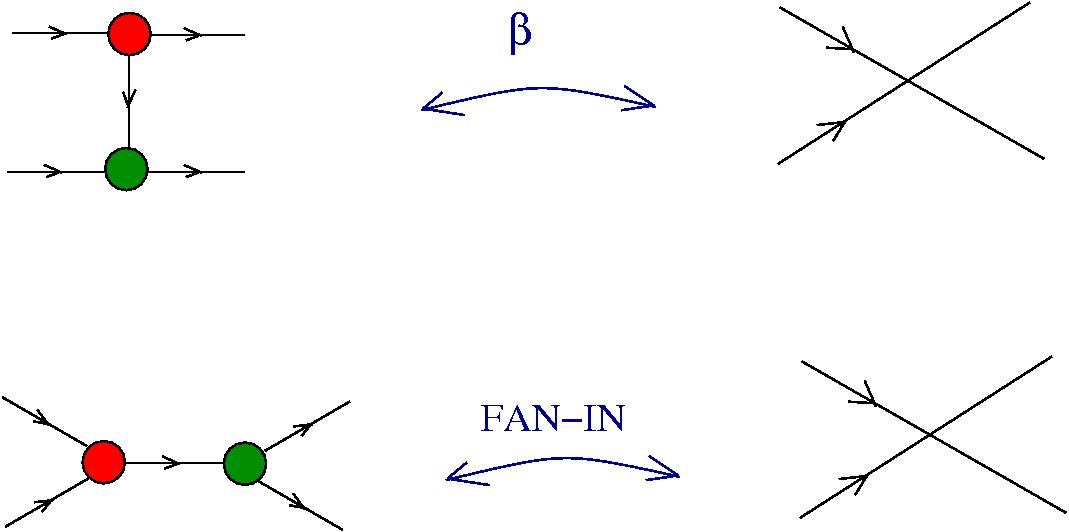}
     \caption{(up) the beta move, (down) the FAN-IN move}
     \label{convention_2}
 
\end{figure}

The list of moves is the following: 
\begin{enumerate}
\item[-] the beta move, Fig.~\ref{convention_2} up,  is the graphic version of beta reduction from lambda calculus. It is a local  graph rewrite version of the Wadsworth \citep{wads} or Lamping \citep{lamping} beta reduction move, in the sense that it can be applied whenever is possible, independently of the fact that the graph, or molecule, represents a lambda calculus term. 
\item[-] the FAN-IN move, Fig.~\ref{convention_2} down, is a dual of the beta move, involving a fanin and fanout node. It is as important as the beta move, being involved into self-multiplication of molecules. 
\end{enumerate}

\begin{figure} [ht]
      
     \includegraphics[width=  75mm]{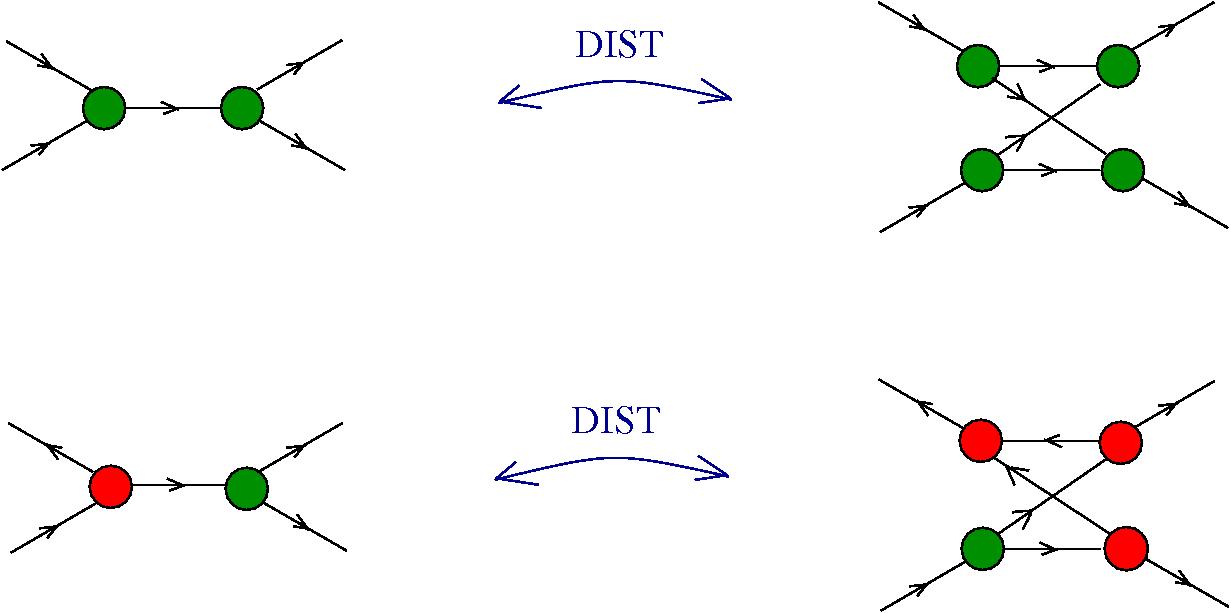}
     \caption{the  DIST moves}
     \label{convention_6}
 
\end{figure}

\begin{enumerate}
\item[-] the DIST moves, Fig.~\ref{convention_6}, provide the mechanism of self-multiplication of molecules, together with the FAN-IN move
\end{enumerate}

\begin{figure} [ht]
      
     \includegraphics[width=  75mm]{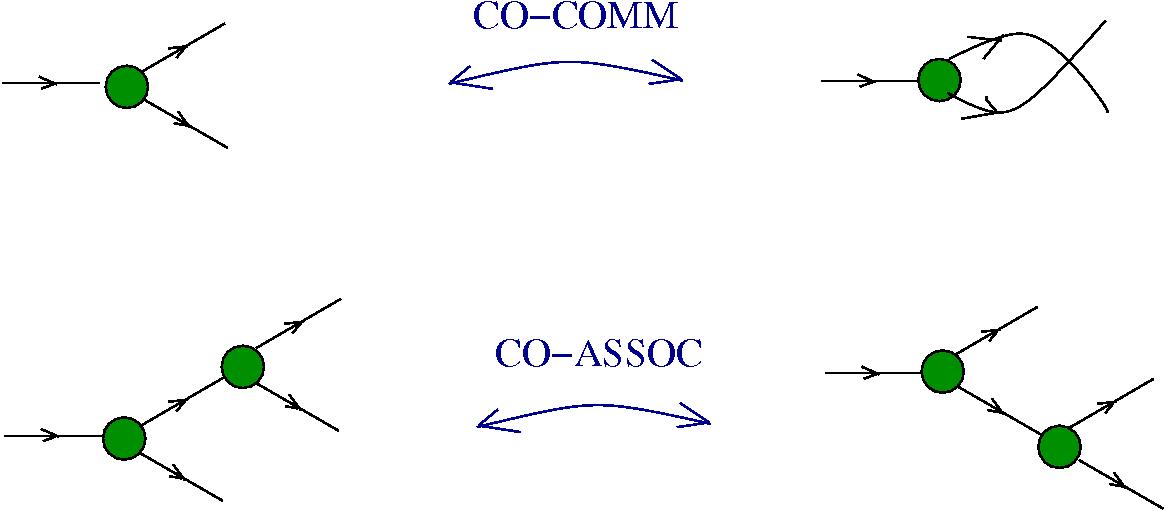}
     \caption{the co-associativity and co-commutativity moves}
     \label{convention_3}
 
\end{figure}

\begin{enumerate}
\item[-] the co-associativity and co-commutativity moves, Fig.~\ref{convention_3}, are a very weak description of the fanout node as a fan-out, in the sense that if we think about the fanout node as being a gate with one input and two outputs which are identical with the input, then graphically such a gate would satisfy the CO-ASSOC and CO-COMM moves. However, the fanout node is not a gate in this formalism, because there is nothing which propagates through the arrows of chemlambda molecules.  
\end{enumerate}

\begin{figure} [ht]
      
     \includegraphics[width=  75mm]{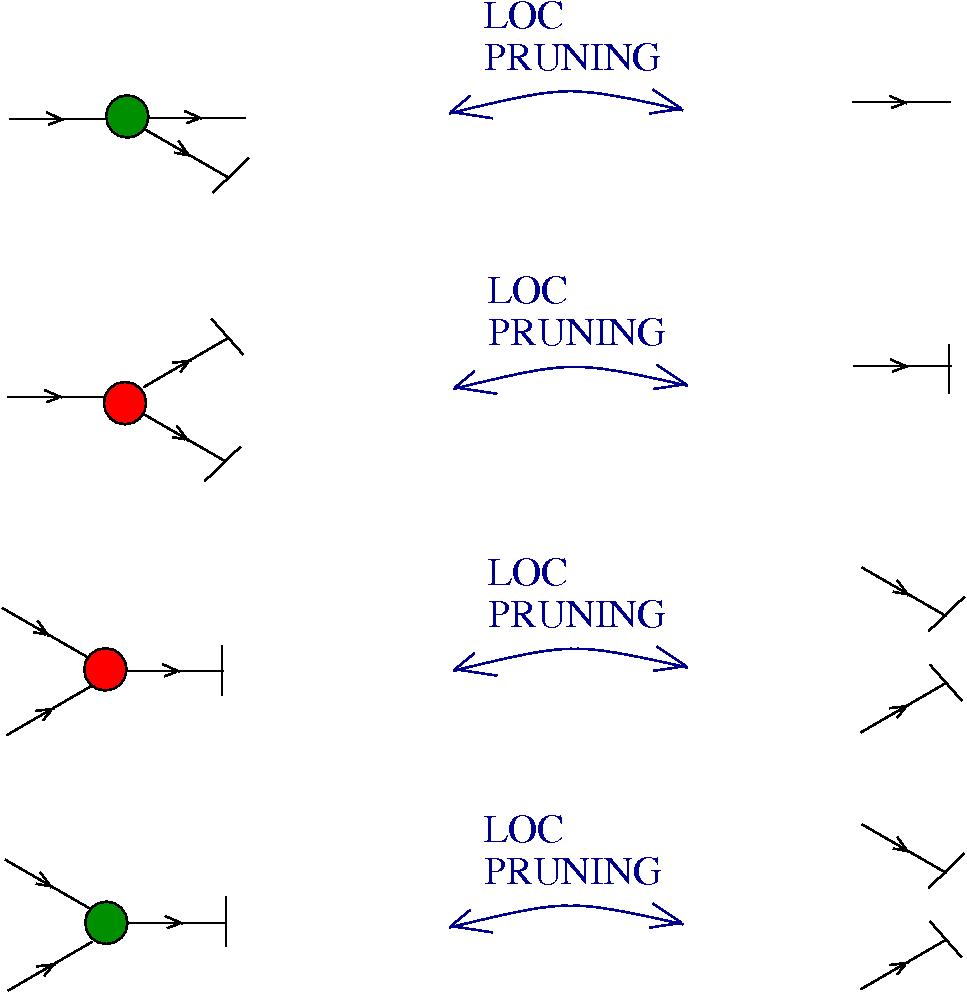}
     \caption{the  local pruning moves}
     \label{convention_4}
 
\end{figure}

\begin{enumerate}
\item[-] the local pruning moves, Fig.~\ref{convention_4}, are useful in both senses, either as moves which destroy the "dead" arrows and nodes, or as moves which enrich the molecule by creating new arrows or nodes.   
\end{enumerate}

\section{Chemlambda and lambda calculus}

\begin{figure} [ht]
      
     \includegraphics[width=  75mm]{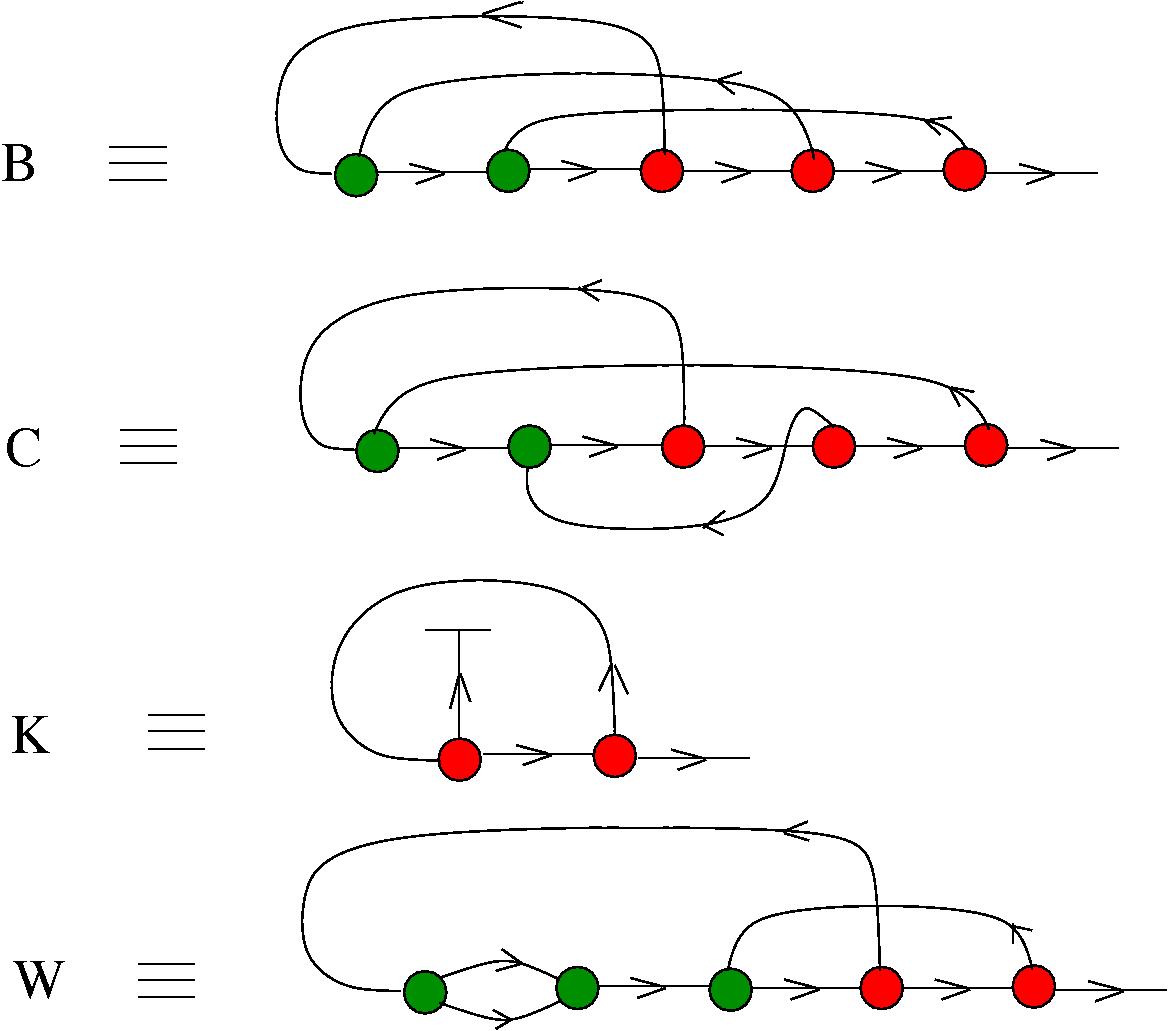}
     \caption{B,C,K,W combinators encoded in chemlambda}
     \label{bckw_2}
 
\end{figure}

Lambda calculus combinators can be encoded as chemlambda molecules. In \citep{buligachem} Theorem 4.2 is given the  encoding of the BCKW system of combinators from Fig.~\ref{bckw_2}. The proof of the theorem has two parts: (a) the reduction relations of the BCKW system can be done in chemlambda, (b) the B,C,K,W combinator molecules {\em can reproduce, or self-multiply}. The conclusion of the theorem is that   {\em chemlambda is Turing universal}.  

We think it is interesting to explain in detail what this self-multiplication means in the chemlambda formalism. 

Remark, after inspection of the Fig.~\ref{bckw_2}, that every combinator molecules has one arrow which points outwards from the molecule, let's call this arrow the "exit arrow".   Recall that we have a fanout node among the basic pieces of chemlambda molecules. In order to prove the Turing universality, we need to be able to transform, by a sequence of chemlambda moves, one combinator molecule with the exit arrow connected to the in arrow of a fanout node, into two copies of the combinator molecule. We call this self-multiplication. (In the GLC formalism this self-multiplication is done via the move  GLOBAL FAN-OUT, but chemlambda has only local moves.)

\begin{figure} [ht]
      
     \includegraphics[width=  75mm]{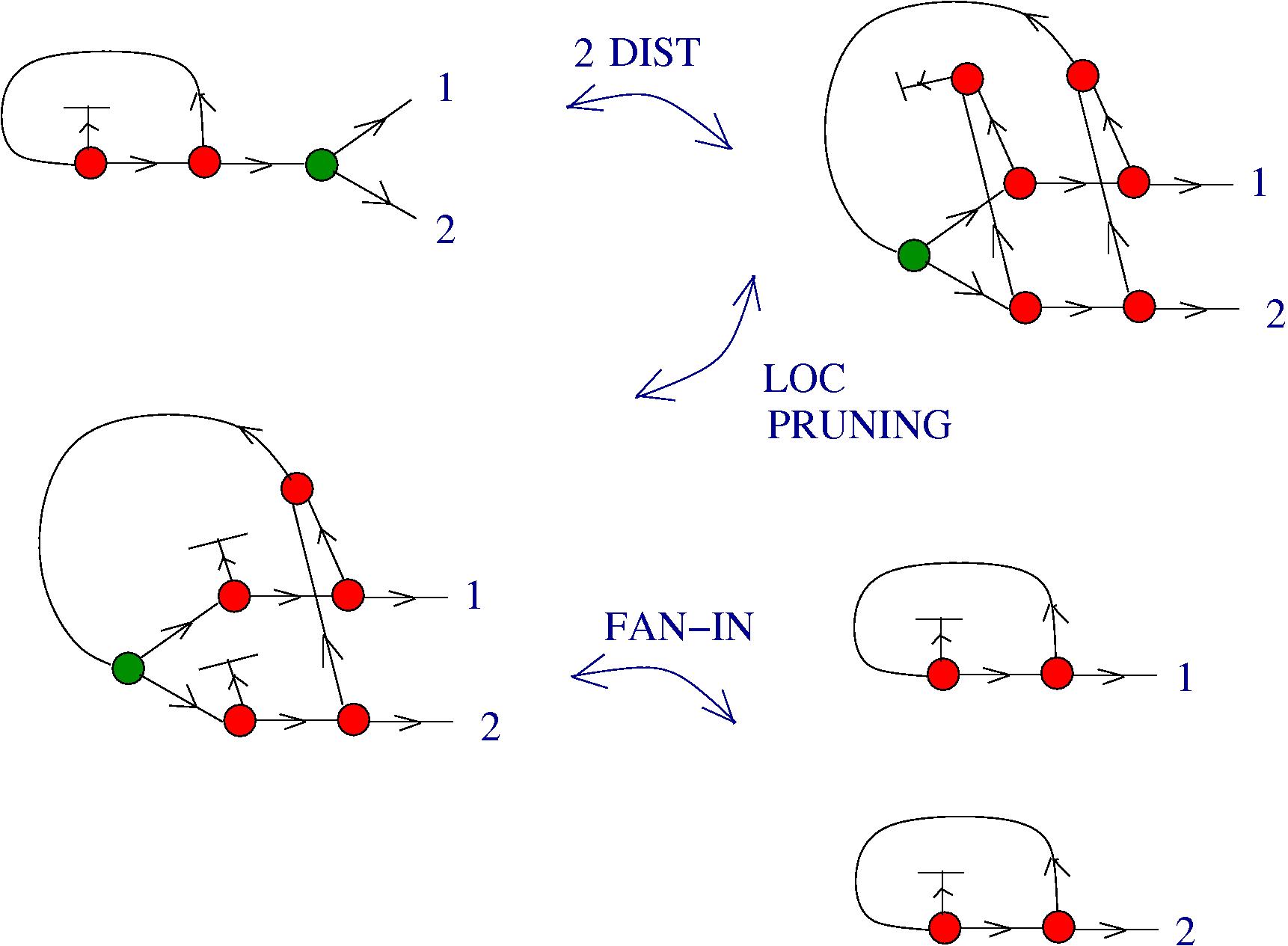}
     \caption{Self-reproduction of the K combinator molecule}
     \label{bckw_6_alife}
 
\end{figure}

As an example, in the  Fig.~\ref{bckw_6_alife} we see how the K combinator molecule self-reproduces, after a string of chemlambda moves.

\section{Propagators, distributors, multipliers and guns}

The self-multiplication of combinator molecules is done by a sequence of local moves of chemlambda. The sequence of moves depends on the combinator molecule. We have seen that self-multiplication is an important ingredient for proving Turing universality of chemlambda. 

Many chemlambda molecules don't encode combinators or lambda calculus terms, moreover, moves like DIST or FAN-IN don't have a clear meaning as seen from the point of view of lambda calculus. The phenomenon of self-multiplication is not restricted to combinator molecules. 

Let us then explore a bit the chemlambda formalism from the point of view of phenomena like self-multiplication, without caring about lambda calculus. 

\begin{figure} [ht]
      
     \includegraphics[width=  75mm]{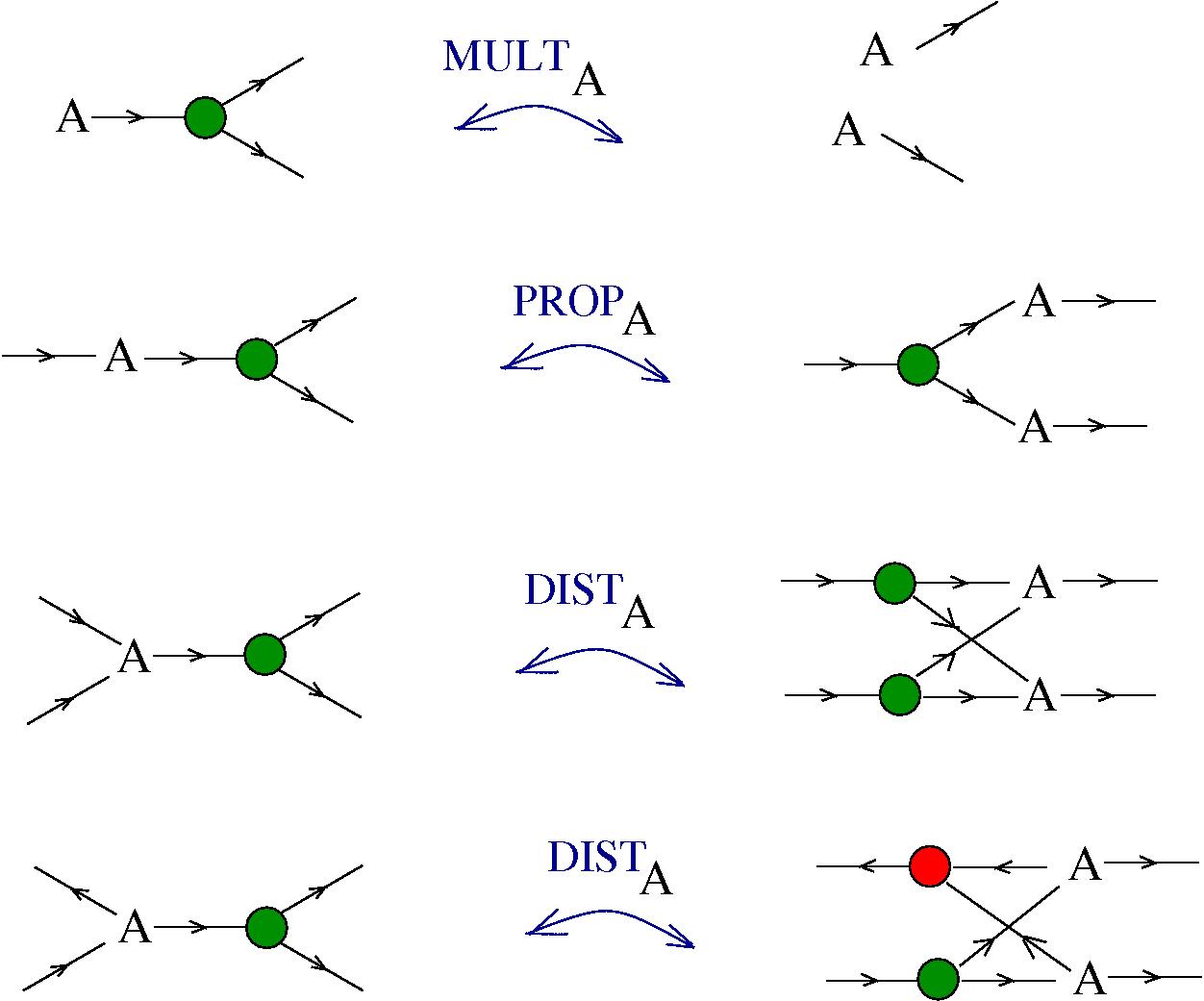}
     \caption{Definition of self-multipliers, propagators, distributors}
     \label{def_multi}
 
\end{figure}

In the Fig.~\ref{def_multi} are defined multipliers, propagators and distributor molecules. A chemlambda molecule with an exit arrow $A\rightarrow$ is a multiplier if there is a sequence of chemlambda moves, denoted by $\displaystyle MULT_{A}$, which produces the self-multiplication of the molecule. For example, any combinator molecule is a multiplier, but there are other multipliers as well. 

A chemlambda molecule $\rightarrow A \rightarrow$ with  distinguished in and out arrows  is a propagator if there is a sequence of chemlambda moves, denoted by $\displaystyle PROP_{A}$, with the effect described in the second row of 
Fig.~\ref{def_multi}. The molecule is called a propagator because it looks like it propagates through the fanout nodes. 

There are two kinds of distributor molecules, described in the 3rd and 4th rows of Fig.~\ref{def_multi}. Compare with the DIST moves from Fig.\ref{convention_6}, which can be interpreted by saying that the application node is a distributor of the first kind and the lambda abstraction node is a distributor of the second kind.

\begin{figure} [ht]
      
     \includegraphics[width=  75mm]{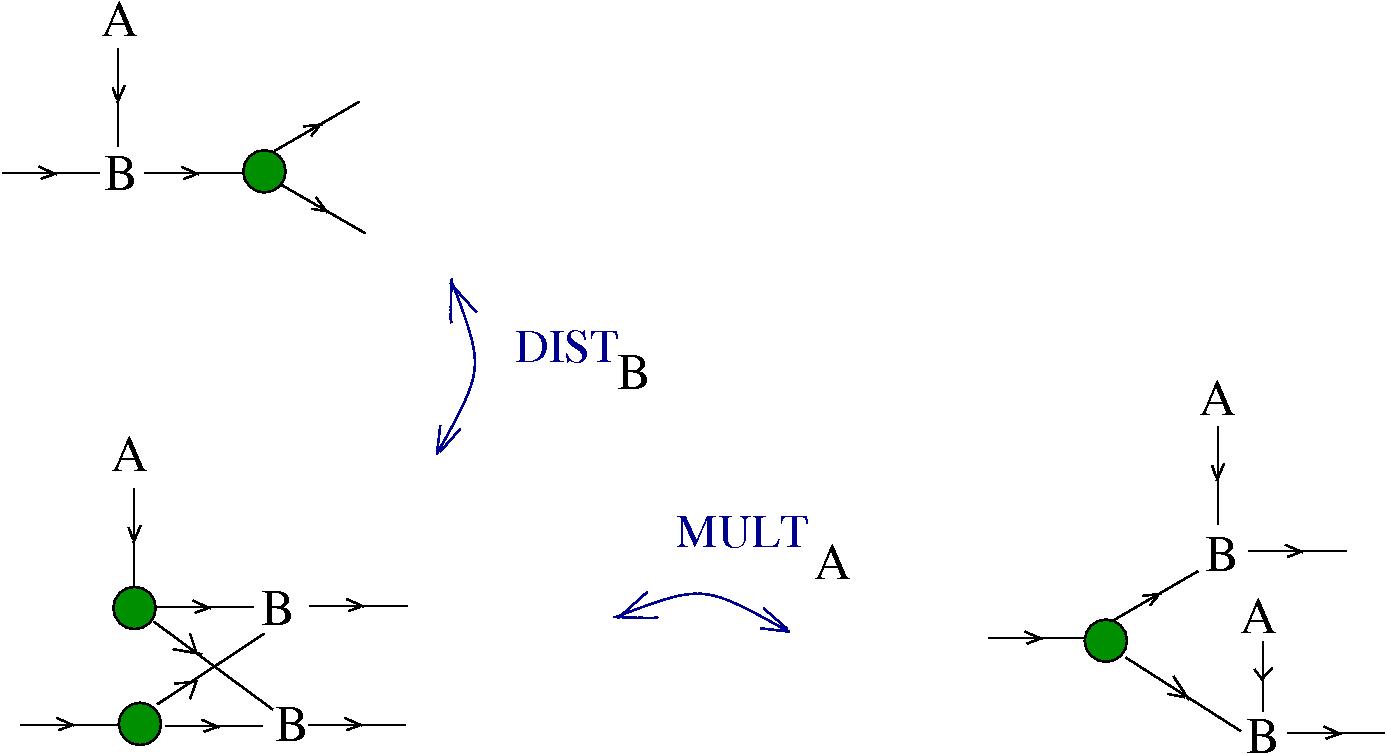}
     \caption{Propagator made from a multiplier and a distributor of the first kind}
     \label{multi_2}
 
\end{figure}

Starting from the mentioned multipliers and distributors, we can make many other interesting molecules. For example, we can make a propagator from a  multiplier $A$ and a distributor of the first kind $B$, as described in Fig.~\ref{multi_2}. 

\begin{figure} [ht]
      
     \includegraphics[width=  75mm]{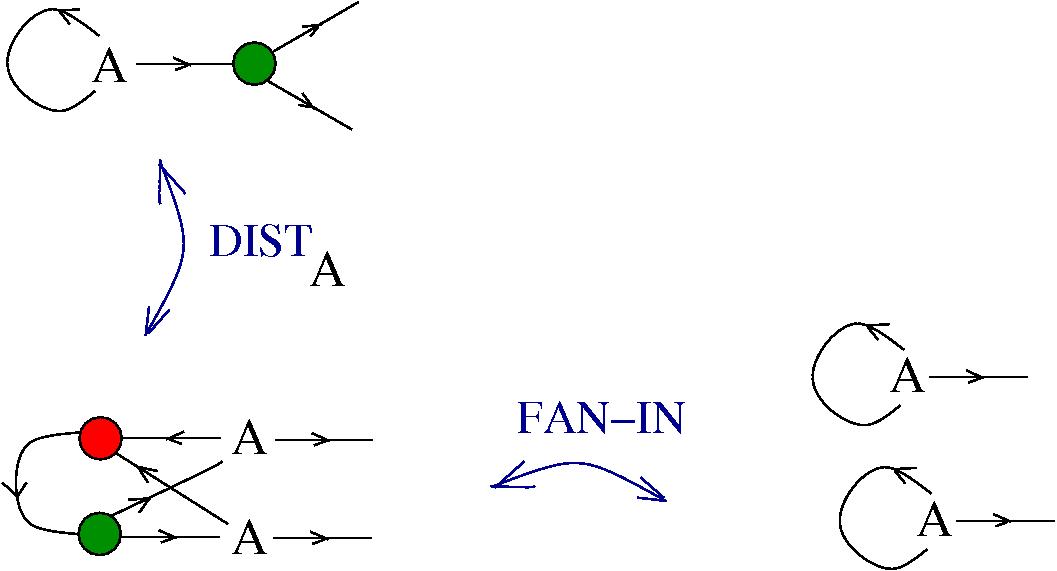}
     \caption{Multiplier made from a distributor of the second kind}
     \label{multi_3}
 
\end{figure}

In Fig~\ref{multi_3} is described a multiplier made from a distributor of the second kind. 

\begin{figure} [ht]
      
     \includegraphics[width=  75mm]{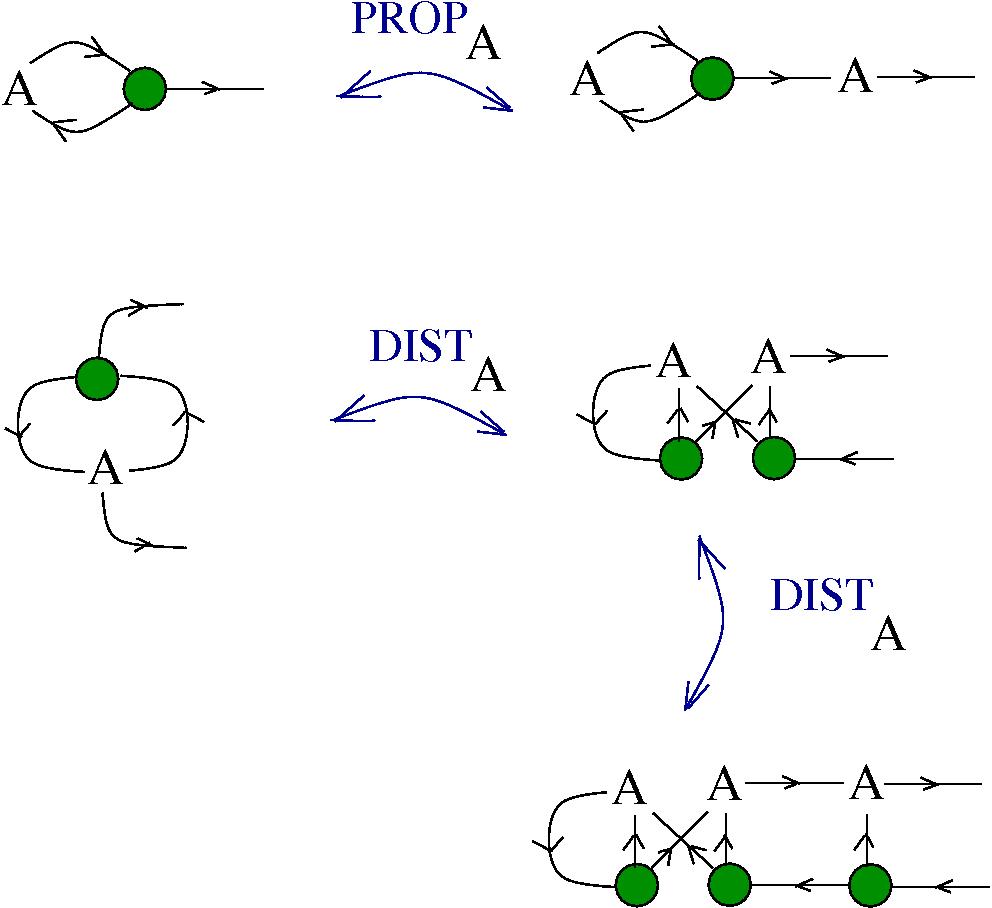}
     \caption{Examples of guns}
     \label{gun}
 
\end{figure}

We can as well make guns, which shoot an endless string of molecules, like in the Fig.~\ref{gun}. On the first row is described a gun made from a propagator molecule and a fanout node. On the second row is described a gun made from a distributor of the first kind and a fanout node. 

 See also \citep{buligachem} Section 3 for other examples of interesting chemlambda molecules, like zippers, sets and pairs.

All these constructions show that we can use chemlambda for building all sorts of synchronous or asynchronous automata (which are not living on a predefined lattice, instead they grow their own lattice). Also, we proved the potential of chemlambda to evolve complex molecules from simple ones. 

\section{The Y combinator and self-multiplication}

In this section we come back to lambda calculus, in order to explain the behaviour of the Y combinator molecule. From the previous sections we learned that self-multiplication is a basic ingredient for encoding the BCKW system of combinators in chemlambda. The moves applied to a combinator molecule represent the reduction of the combinator. Self-multiplication is needed in order to produce copies of a part of the combinator molecule, with the purpose of further reducing one of the copies, while having at our disposal the other copy for further needs. 

Seen like this, it seems that self-multiplication is also a basic ingredient for recursion. In lambda calculus there is the iconic Y combinator which represents the essence of recursion. In the following we shall see that, however, 
self-multiplication is not directly needed in the reduction of the Y combinator. 
\begin{figure} [ht]
      
     \includegraphics[width=  85mm]{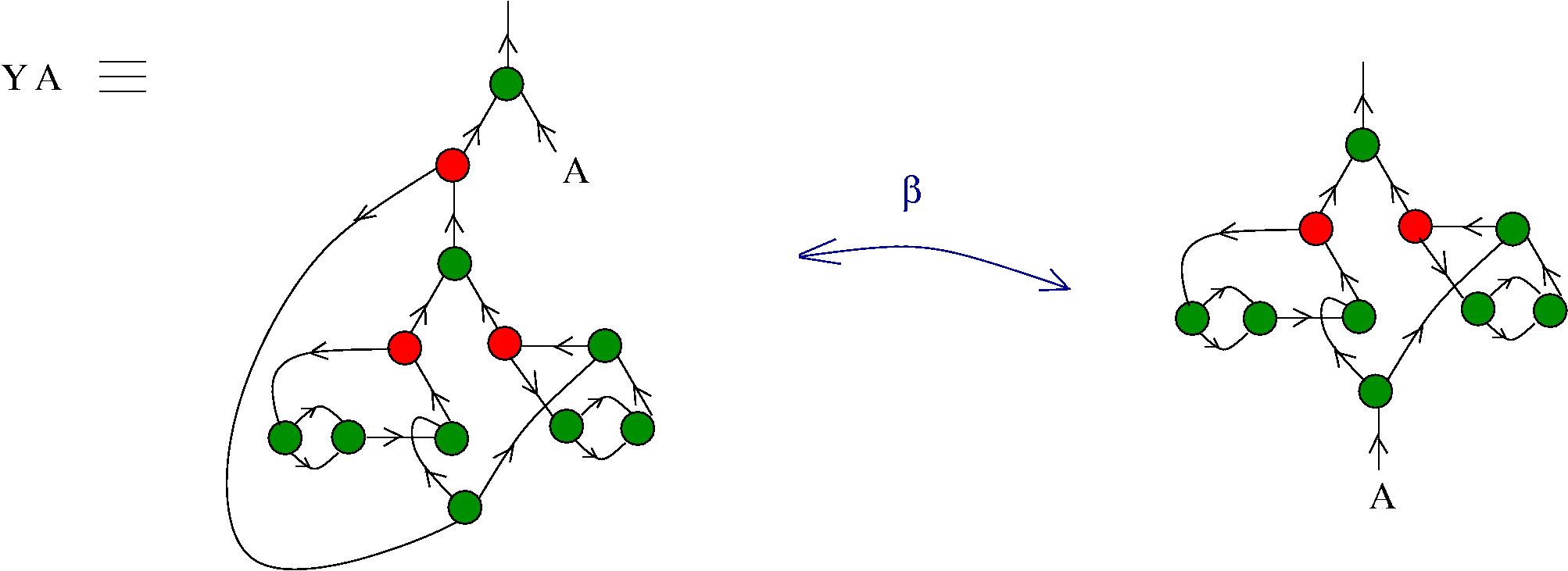}
     \caption{the $YA$ combinator molecule and a first beta move}
     \label{ya}
 
\end{figure}

\begin{figure} [ht]
      
     \includegraphics[width=  80mm]{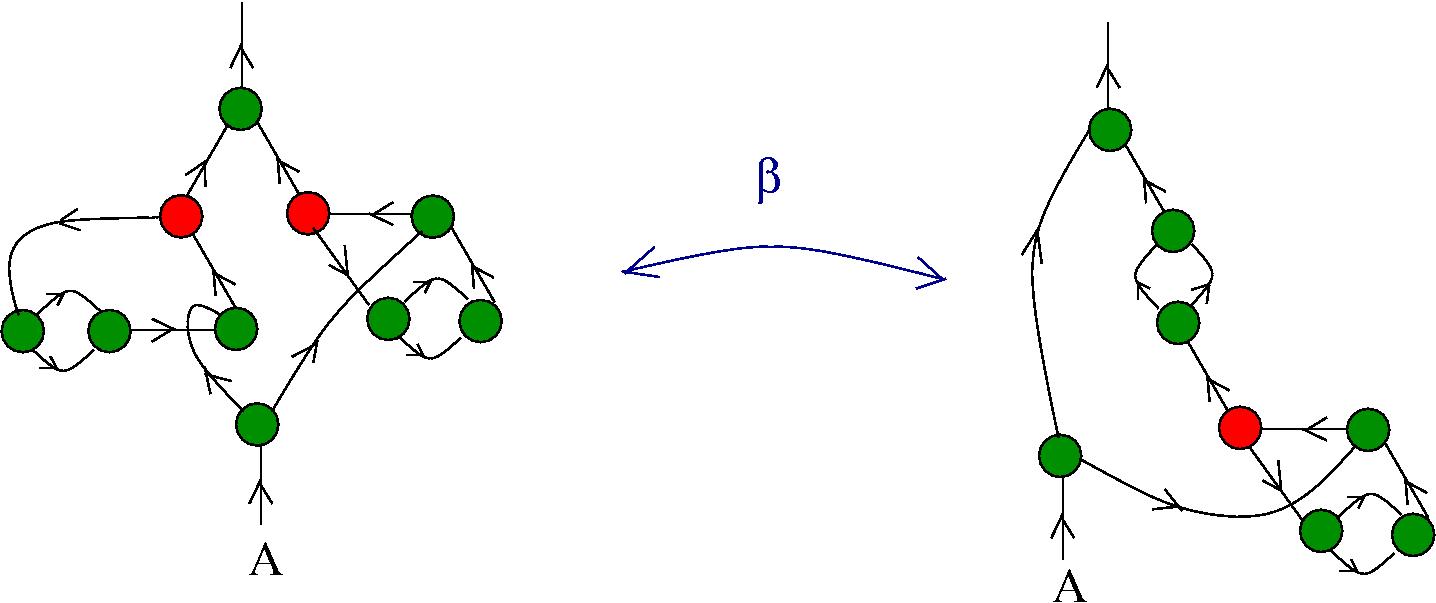}
     \caption{second beta move applied to the $YA$ molecule}
     \label{ycombi_s_1}
 
\end{figure}

\begin{figure} [ht]
      
     \includegraphics[width=  80mm]{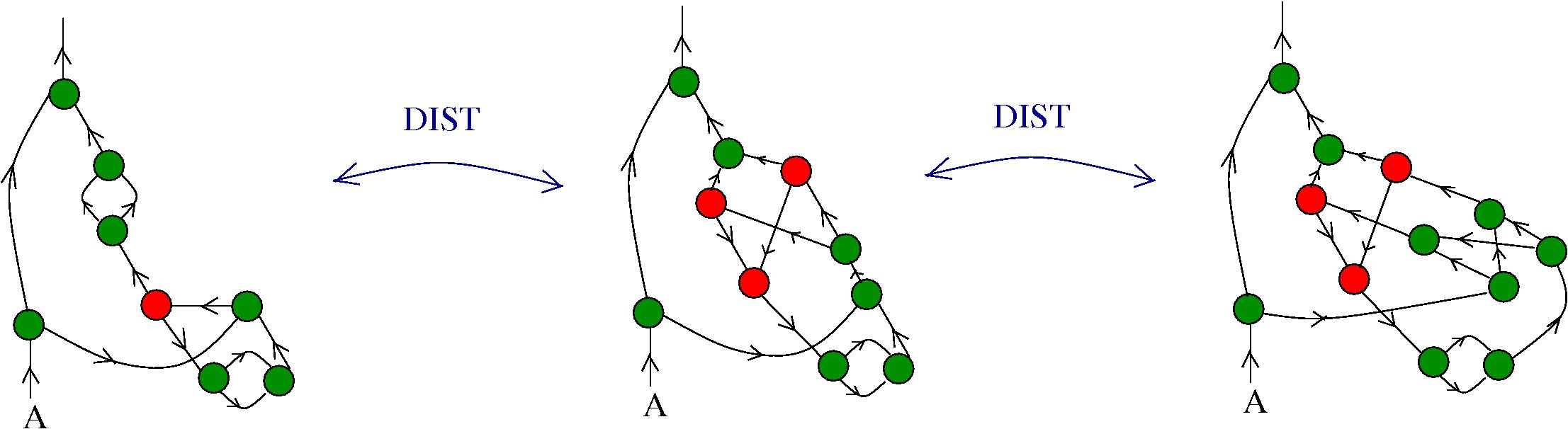}
     \caption{next step of reduction, two DIST moves}
     \label{ycombi_s_2}
 
\end{figure}

The Y combinator has the expression 
$$Y = \lambda y.(\lambda x. y(xx)) (\lambda x. y(xx))$$
and it has the following property: for any lambda term $A$ the expression $YA$ reduces to $A(YA)$. In particular, if $A$ is another combinator, then $YA$ is a fixed-point combinator for $A$.

In lambda calculus the string of reductions is the following sequence of beta moves: 
$$YA \rightarrow (\lambda x. A(xx)) (\lambda x. A(xx)) \rightarrow $$ 
$$ \rightarrow  A ( (\lambda x. A(xx)) (\lambda x. A(xx)) ) = A(YA)$$
We see that the during the reduction process we needed a multiplication of the combinator $A$. 

Let us pass to the chemlambda encoding of the Y combinator. With $A$ another combinator molecule, the combinator molecule which encodes  $YA$ is the one from the left hand side of the Fig.~\ref{ya}.

After the application of a beta move, it transforms into the molecule from the right hand side of Fig.~\ref{ya}. Continuing from the Fig.~\ref{ya}, there is a second beta move which can be applied, as in Fig.~\ref{ycombi_s_1}.

There are two DIST moves, one of the first kind, the other of the second kind, which are applied, as in Fig.~\ref{ycombi_s_2}.

\begin{figure} [ht]
      
     \includegraphics[width=  80mm]{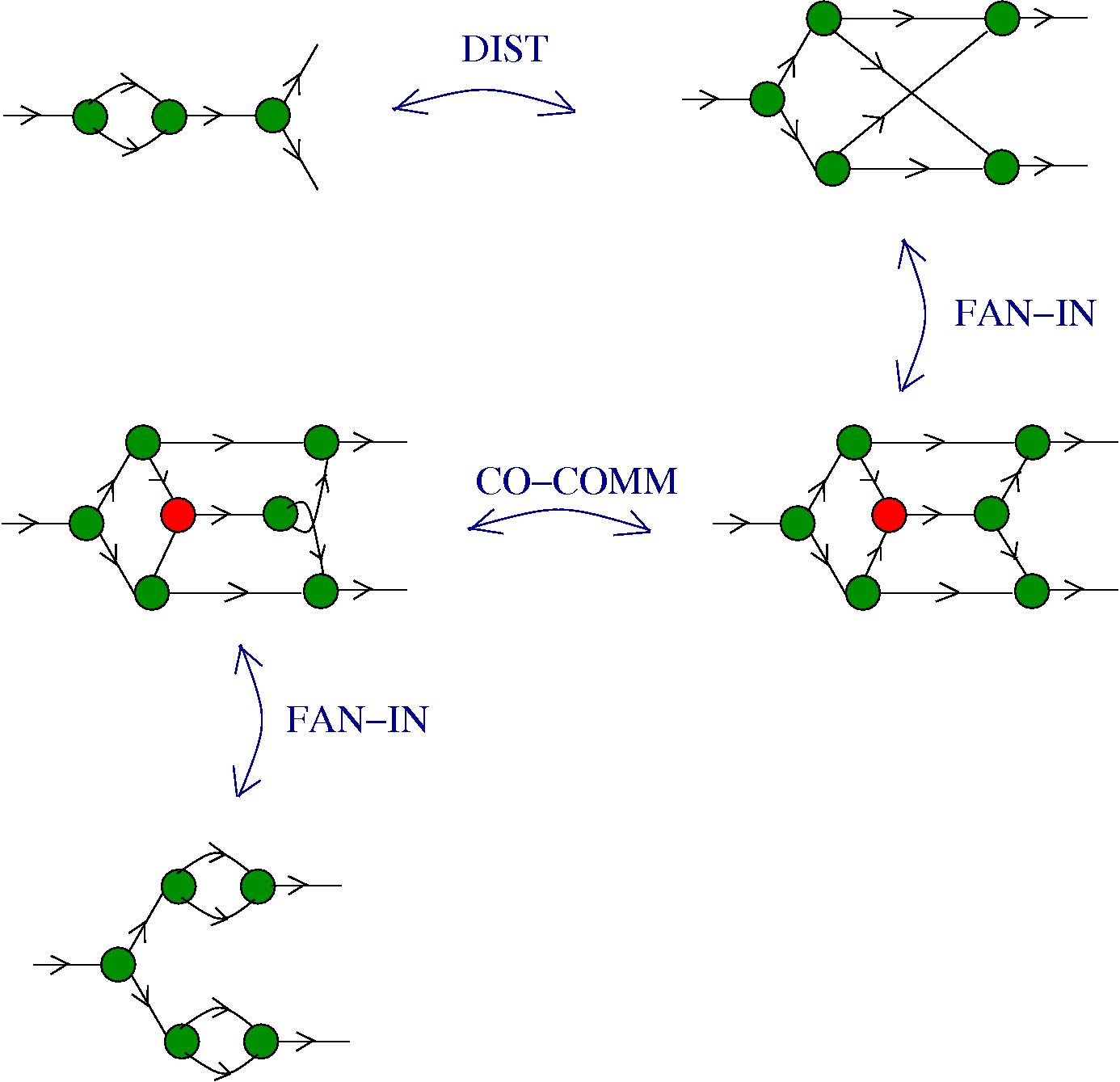}
     \caption{the bit is a propagator}
     \label{bit}
 
\end{figure}

Let's see how we can reduce further this molecule, until we obtain one which corresponds to $A(YA)$. We shall use  the fact that a certain molecule, called  {\em the bit} is a propagator, as proved in Fig.~\ref{bit}. The bit molecule  corresponds to the expression $(xx)$ which appears repeatedly in the $Y$ combinator. 

\begin{figure} [ht]
      
     \includegraphics[width=  80mm]{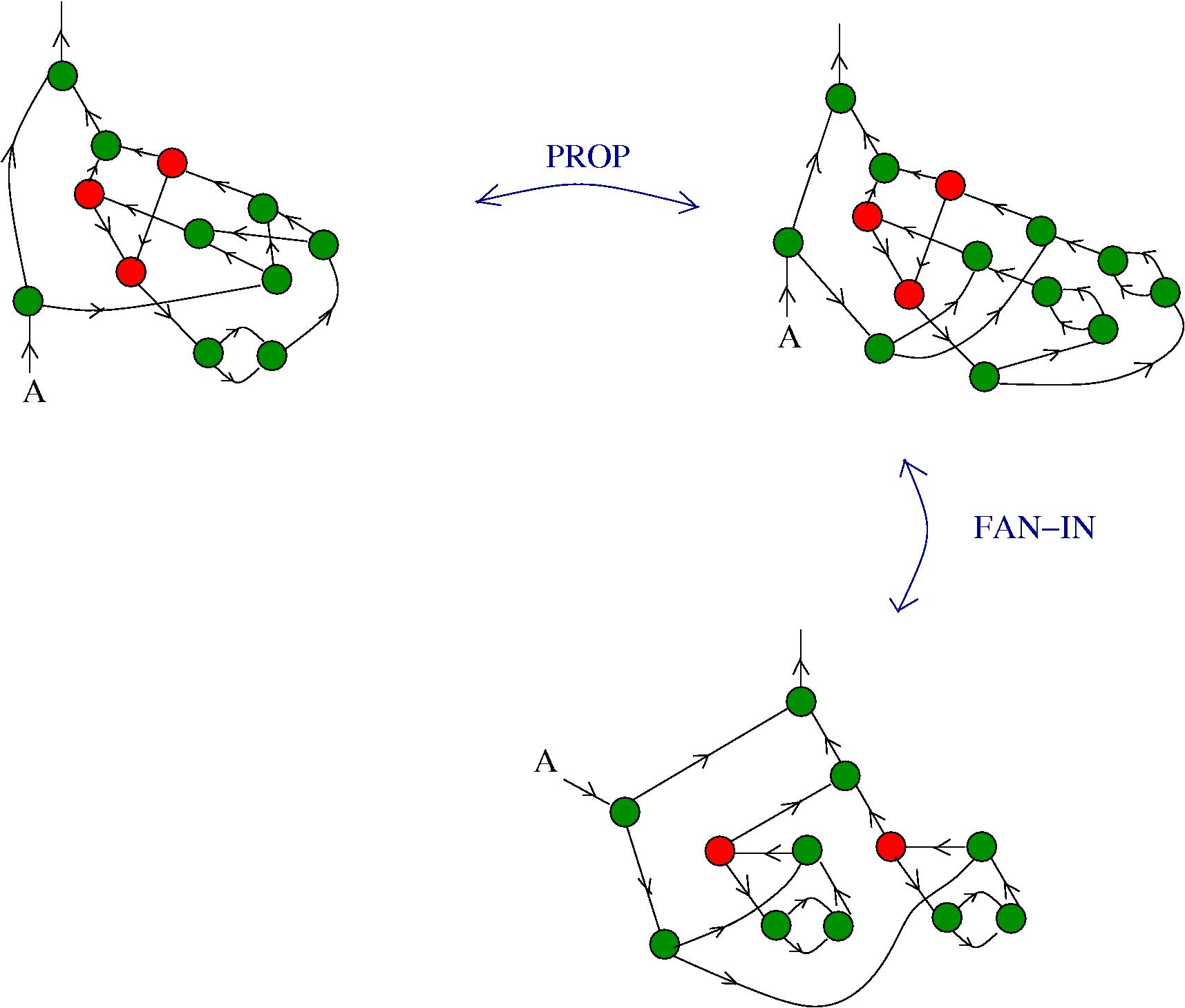}
     \caption{last two moves of the reduction of $YA$ to $A(YA)$}
     \label{ycombi_s_3}
 
\end{figure}

We continue from the Fig.~\ref{ycombi_s_2} and we apply the PROP move of the bit and then a FAN-IN move, as in the Fig.~\ref{ycombi_s_3}. 

The last molecule corresponds to $A(YA)$, if we interpret the fanout nodes as real fan-out gates. 

\begin{figure} [ht]
      
     \includegraphics[width=  80mm]{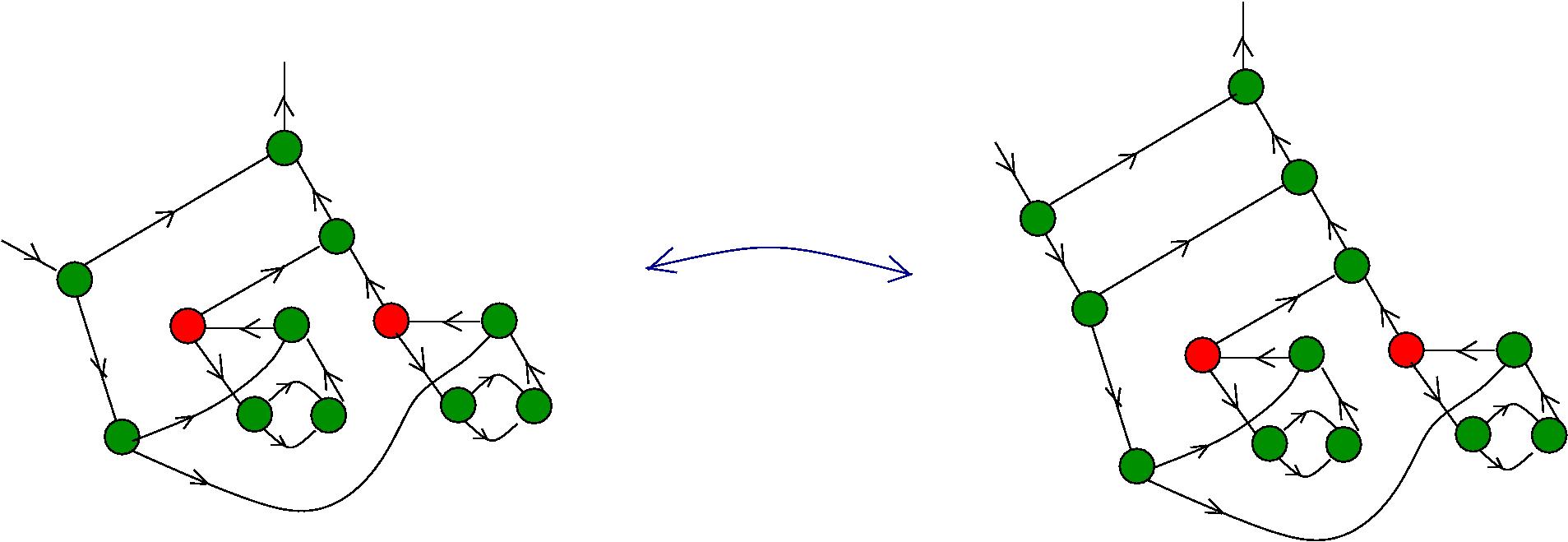}
     \caption{the $Y$ molecule is a gun}
     \label{ycombi_s_4}
 
\end{figure}

Surprisingly, during the reduction there was no need to use the fact that the combinator molecule $A$ is a multiplier! This shows that the $Y$ combinator molecule can be used as a fixed point combinator with {\em any} other chemlambda molecule. That is because the $Y$ combinator molecule is a gun which shoots fanout nodes, Fig.~\ref{ycombi_s_4}. 

\section{A topological version of chemlambda}
\label{tglc}

In \citep{Web} Section 5 is proposed a topological version of GLC, called TGLC. We can do the same with chemlambda. The idea is that we may imagine formalisms which are equivalent with  GLC and chemlambda, even if visually they seem  different. 
 
\begin{figure}[ht]
      
     \includegraphics[width=7cm]{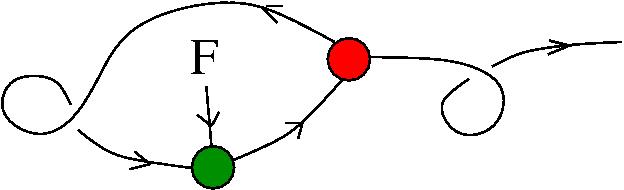}
     \caption{Topological Fixed Point Combinator}
     \label{tfix}
 \end{figure}

\begin{figure}[ht]
      
     \includegraphics[width=8cm]{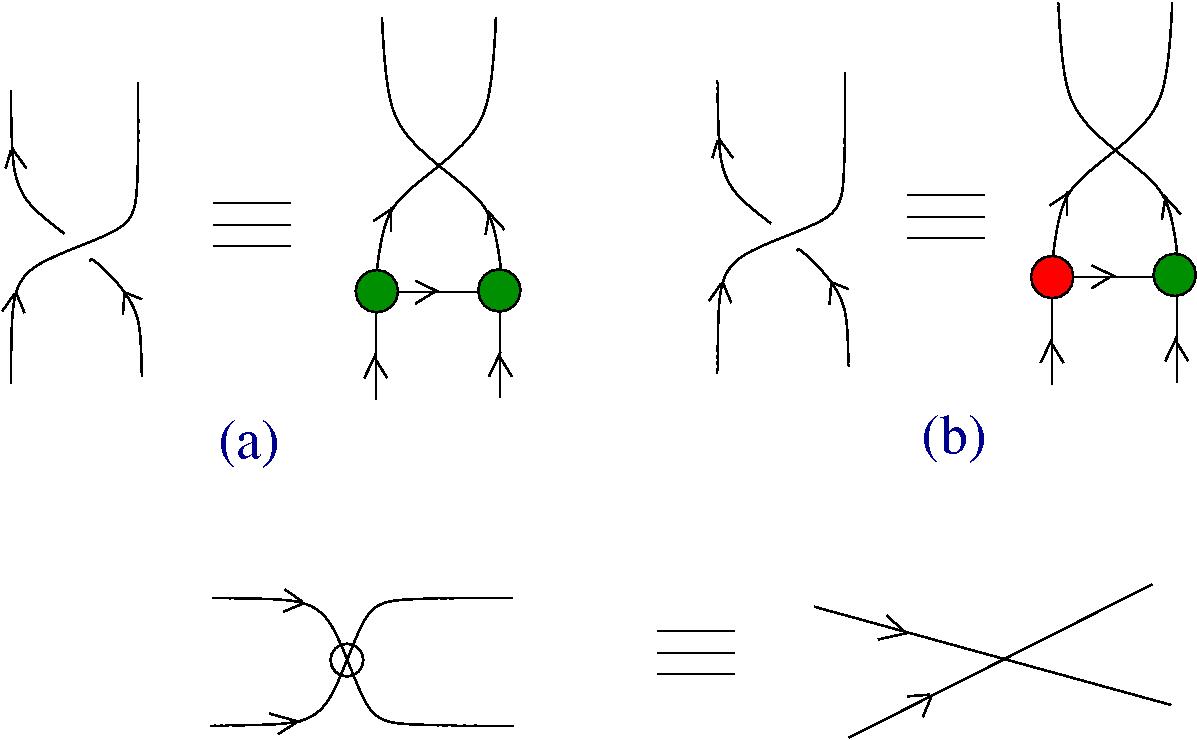}
     \caption{first row, two possible translations from crossings to chemlambda, second row a virtual crossing}
     \label{twoknots}
 
\end{figure}

For a topological version of chemlambda we may use some of the basic nodes of chemlambda together with  knot diagrams crossings. In Fig.~\ref{twoknots} we give two possible translations of crossings into chemlambda: (a) as a pair of a fanout and application node, corresponding to the proposal made in \citep{Web} Section 5, or (b) as a pair of a lambda abstraction node and an application node, as proposed in \citep{bgraph} Section 6.  A crossing is a 4 valent vertex. {\it Virtual crossings}, i.e. encircled crossings of graphical lines, may be used for making our graphs globally planars instead of only locally planar, as previously. 

If we stick to the choice (a) then we obtain a topological version of chemlambda, that
has the form of knot diagrams equipped with extra lambda nodes and multiplication nodes.

In Fig.~\ref{tfix} we illustrate the basic fixed point combinator 
$$G = \lambda x.F(xx)  \lambda x.F(xx)$$
 In this knot diagrammatic convention,  the two self-multiplications that occur at two levels in this expression are instantiated by the two curls in the graph.

\begin{figure}[ht]
      
     \includegraphics[width=7cm]{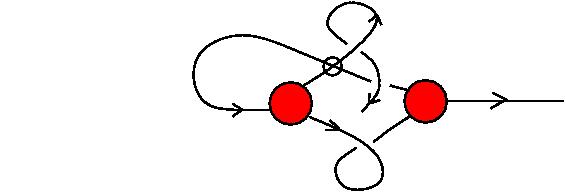}
     \caption{Topological Y - Combinator}
     \label{yfix}
 
\end{figure}

Similarly, in Fig.~\ref{yfix} we illustrate a topological expression for the $Y$-combinator.

Note how the structure of this combinator takes on the hybrid nature of tangle diagram infused with curls and lambda nodes. It is natural to use virtual crossings in graph theory and in fact there is an extension of knot theory  that allows exactly such virtual crossings in the knot diagrams. 

\begin{figure}[ht]
\includegraphics[width=8cm]{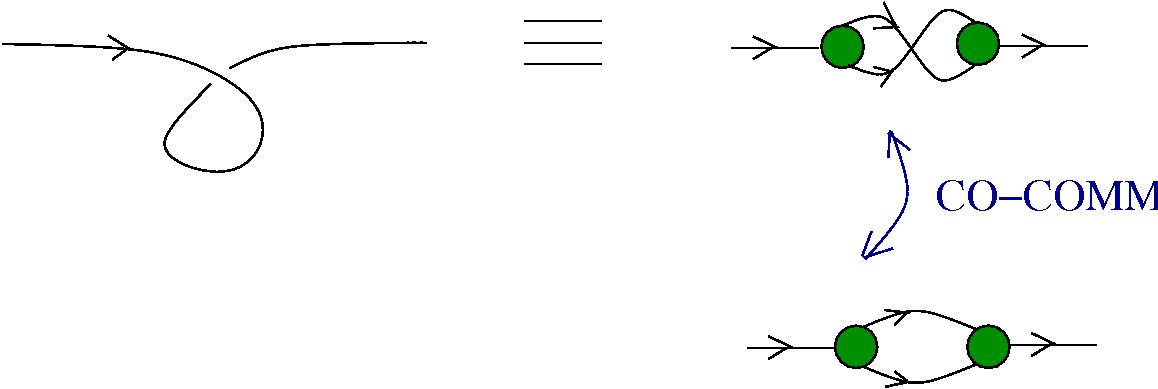}
\caption{a curl is a bit}
\label{curl_bit}
\end{figure}

In Fig.~\ref{curl_bit} we see that, via a CO-COMM move, a curl is a bit, the molecule which appears in Fig~\ref{bit}.

The fact that alpha reduction is not needed in  chemlambda  
 due to the absence of variables and the presence of direct
connections that effect interactions is part of a link of this formalism
with the formalisms at the knot theoretic  and topological level.       
One  difference between knot theoretic considerations and lambda
calculus considerations is in the fact that we do not usually think of a
knot diagram as a computing element that undergoes moves and reductions
for the sake of a computation. But this is not always so.
For example, the skein algorithms such as the bracket polynomial algorithm
can be regarded as a reduction process that produces two new graphs from
each crossing in the knot diagram. This is similar to allowing free beta
reduction in the lambda calculus graphs. What must be done however in the
knot theoretic case is to collect up all the end calculation results and
add them together. This is what is meant by a formula like
$$\langle K \rangle = \Sigma_{S} \langle K|S \rangle.$$
(See \citep{kauffman4}.) Each $S$ is a pattern of reductions leading to a specific algebraic value
$\langle K|S \rangle .$ The topological invariance occurs at the level of the sum of all of
these contributions. 

An analogous situation could occur in a stochastic version of chemlambda, where one would need the average over all the results of the many branching graph rewrites.

\footnotesize
\bibliographystyle{apalike}
\bibliography{alife_glc_cut}

\end{document}